\documentclass[twocolumn]{IEEEtran}
\usepackage{graphicx}
\usepackage{amssymb}
\usepackage{physics}
\usepackage{cite}
\usepackage{amsmath}
\usepackage{algorithmic}
\usepackage{color}
\usepackage[table,xcdraw]{xcolor}
\usepackage{multirow}
\usepackage{soul}

\newtheorem{lemma}{Lemma}

\begin{document}
\title{Deep Global-Connected Net With The Generalized Multi-Piecewise ReLU Activation in Deep Learning}
\author{Zhi Chen, Pin-Han Ho
\thanks{Z. Chen and P.-H. Ho are with the Department of Electrical and Computer Engineering,
University of Waterloo, Waterloo, Ontario, N2L 3G1, Canada (e-mail: {z335chen, p4ho}@uwaterloo.ca).}}

\maketitle
\begin{abstract}
Recent Progress has shown that exploitation of hidden layer neurons in convolutional neural networks (CNN) incorporating with a carefully designed activation function can yield better classification results in the field of computer vision. The paper firstly introduces a novel deep learning (DL) architecture aiming to mitigate the gradient-vanishing problem, in which the earlier hidden layer neurons could be directly connected with the last hidden layer and feed into the softmax layer for classification. We then design a generalized linear rectifier function as the activation function that can approximate arbitrary complex functions via training of the parameters. We will show that our design can achieve similar performance in a number of object recognition and video action benchmark tasks,
such as MNIST, CIFAR-10/100, SVHN and UCF YoutTube Action Video datasets,
under significantly less number of parameters and shallower network infrastructure, which is not only promising in training in terms of computation burden and memory usage, but is also applicable to low-computation, low-memory mobile scenarios.
\end{abstract}


\IEEEpeerreviewmaketitle

\section{Introduction}
Since Hinton et al applied a deep convolution neural network (CNN) to achieve great success in ImageNet competition in 2012 \cite{Hinton'ImageNet-2012}, CNN has been well recognized as a powerful tool for the computer vision applications in the recent years. This, however, cannot be possible without the availability of massive image/video datasets collected in the internet-of-things era, as well as the innovation of high-performance parallel computing resources. Exploiting these resources, novel ideas, algorithms as well as modification on the network architecture of deep CNN has been experimented to achieve higher performance in different computer vision tasks. Deep CNNs thus were found to be able to extract rich hierarchal features from raw pixel values and achieved amazing performance for classification and segmentation tasks in computer vision \cite{Hinton'ImageNet-2012}-\cite{Lee-ICML-16}.

Starting with LeNet-5 \cite{LeCun-MNIST'98}, a typical CNN usually consisting of several cascaded convolutional layers, optionally pooling layers (average pooling or max pooling), nonlinear activations as well as fully-connected layers, followed by a final softmax layer for classification/detection tasks, where the convolution layer is employed to learn the spatially local-connectivity of input data for feature extraction, pooling layer is for reduction of receptive field and hence overfitting, and nonlinear activations for boosting of learned features. Since then, Deeper (more layers) and wider (larger layer size) variants of the standard CNN architecture has been introduced and experimented on various computer vision datasets, which are shown to achieve state-of-the-art performance, compared with other machine learning techniques. For example, the elegant GoogleNet \cite{Szegedy-Deeper'14} achieved the best performance on ILSVRC 2014 competition, which employed $5$ million parameters and $22$ layers consisting of inception modules. To avoid overfitting for deep neural networks, some regularization methods
are invented, such as dropout \cite{Hinton'14-Dropout'JMLR} or dropconnect \cite{DropConnect'13}, which
turns off the neurons learned with a certain probability in training, and hence prevents
the co-adaptation of neurons during the training phase. In \cite{Szegedy'-BN-15} and \cite{Salimans-Weight-Normalization-16},
the authors proposed batch normalization as well as weight normalization methods, respectively, providing powerful ways
to allow higher learning rates and be less dependable on careful initialization for deep neural networks. These techniques are now commonly employed in
modern deep CNN architectures to avoid overfitting.

On the other hand, other than the deeper architecture, one key feature of the success of deep CNN architecture is the use of appropriate nonlinear activation functions that define the value transformation from the input to output. It was found that the linear rectifier activation function (ReLU) \cite{Bengio-sparse-ReLU'11} can greatly boost performance of CNN in achieving higher accuracy and faster convergence speed, in contrast to its saturated counterpart functions, i.e., sigmoid and tanh functions. ReLU only applies identity mapping on the positive side while drops the negative input, allowing efficient gradient propagation in training. Its simple functionality enables training on deep neural networks without the requirement of unsupervised pre-training and paved the way for implementations of very deep neural networks. On the other hand, one of the main drawbacks of ReLU is that the negative part of the input is simple dropped and are not updated in training in backward pass, causing the problem of dead neurons which may never be reactivated again and potentially results in lost feature information through the back-propagation.
To alleviate this problem, some new types of activation functions based on ReLU are reported for CNNs. Ng et al \cite{Maas'13-ReLU} introduced the Leaky ReLU assigning a non-zero slope to the negative part, which however is a fixed parameter and not updated in learning. Kaiming et al pushed it further to allow the slope on the negative side to be a learnable parameter, which is hence named Parameter ReLU (PReLU) in \cite{Kaiming'relu'15}. Further, \cite{Xu'13-S-shaped} introduced an S-shaped ReLU function. In \cite{'Clevert'Elu'2015}, the authors introduced the ELU activation function, which assigns an exponential function on the negative side to zeroing the mean activation. The network performance is improved at the cost of the increasing computation burden on the negative side
(as an exponential function is employed for activation in the negative part), compared with other variants of ReLU. Nevertheless, all of these functions lack the ability to mimic complex functions on both sides in order to extract necessary information relayed to the next level. In addition, Goodfellow et al introduced a maxout function which selects the maximum among $k$ linear functions for each neuron as the output in \cite{Goodfellow-Maxout'13}. While maxout has the potential to mimic complex functions and perform well in practice, it takes much more parameters than necessary for training and thus reduces its popularity in terms of computation and memory usage in real-time and mobile applications.

The other design aspect of deep CNN is on the size of the experimented network and further on the interconnection architecture of different layers. In fact, network size has a dominant impact on the performance of the neural network,  and a natural way for performance improvement is to simply increase its size, by either depth (number of layers) or width (number of units in each layer). This works well suited to the case with a massive number of labeled training data. However, when the amount of labeled training data is small, this potentially leads to overfitting and works poorly in the inference stage for unseen unlabeled data. On the other hand, a large-size neural network requires large amounts of computing resources for training, and a non-necessary large size only ends up with the waste of such valuable resources, as most learned parameters may finally
found to be/near zero and can be simply dropped.
Therefore, instead of changing network size, there is an emerging trend in exploiting the interconnection of the network, i.e., making better use of features learned at the hidden layers in contrast to conventional cascaded structure, to achieve better performance. The intuition follows from the fact that, the learned gradients flow from the output layer to the input layer, they could easily be diluted or even vanished when it reaches the beginning of the network, and vice versa, which hence greatly prohibits performance improvement in the previous decades. In fact, as discovered in the literature, addressing these issues will allow the same performance achievable with smaller network size and smaller amount of parameters, and thus makes real-time mobile/embedded applications practical. On the other hand, addressing these issues also makes very-deep networks trainable,
as the information of the hidden layers can be efficiently flowed in both forward and backward stages.
Therefore, some research works hence have started to tackle this problem in recent years.
In \cite{Szegedy-Deeper'14}, an inception module concatenating the feature maps produced by filters of different sizes, i.e., a wider network consisting of many parallel convolutional networks with filters of different sizes, was proposed to improve the network capacity as well as the performance. In \cite{He'ResNet'16}, the authors proposed a residual-network architecture (ResNet) to ease the training of networks, where higher layers only need to learn a residual function with respect to the features in the lower layers. In this way, every node only needs to learn the residual information transferred from lower layers and thus is expected to achieve a better local optimal point with better performance and less convergence time.
With the residual module, the authors found that very-deep network (e.g., a $160$-layer ResNet was successfully implemented) is trainable and can achieve amazing performance
than its counterparts in the literature. In \cite{Komodakis-Wide-Residual-Net'16}, a wide residual network is proposed,
which decreases depth but increase width of he ResNet module and also achieves the state-of-art performance in experimentation.
In \cite{Stochastic-Resnet'16}, the authors proposed a stochastic depth approach by training short networks and using deep networks at test time. This is achieved by
allowing a subset of network modules simply dropped and replaced by identity functions in training. In doing so, the authors discovered that
training very-deep networks beyond $1020$-layer network is feasible
and performs extremely well.
In \cite{highway'15}, a highway network is proposed by the use of gating units
which learn to regulate the flow of information through the network, and achieves the state-of-art performance.

Further, \cite{M'13-NIN} introduces a Network in Network (NIN) architecture that contains several micro multi-layer perceptrons between the convolutional layers to exploit complicated features of the input information. By fine tuning the learned features using the micro multi-layer perceptrons
between convolution layers,  the NIN network architecture greatly
reduces the number of parameters while still achieving great performance in the public datasets.
In \cite{Lee'DeepSuper'14}, the authors introduced a new type of regularization, where auxiliary classifiers are used on the hidden layers with the main classifier of the output layer, so as to strengthen the features
learned by hidden layers. This is somewhat similar to the layer-by-layer pre-trained
method in CNN in which the heavy supervised learning process may cause the risk of model overfitting. Further, the use of the auxiliary classifiers introduces a lot of parameters, however only the main classifier is employed in the inference stage.

In \cite{Deep-fused'16}, the authors introduced deeply-fused nets to improve the end-to-end information flow by combining many intermediate layers of different base networks. Further, in \cite{Densely-connected'16}, the authors introduced a densely connected architecture within every block to ensure high information flow among layers in the network, and achieved great success in
performance evaluations in most public datasets.  In \cite{Stochastic-pooling'13} and \cite{Pyramid-pooling'14},
the stochastic pooling and pyramid pooling strategies were presented to boost vision recognition performance.

Still, deep CNN is subject to some open problems. One is that the features learned at a intermediate hidden layer could be lost at the last stage of  the classifier after passing through many later layers. Another is the gradient vanishing problem, which could cause training difficulty or even infeasibility. They are hence receiving increasing interest in the literature and the industry.
In this paper, we are also motivated to mitigate such obstacles by targeting at the tasks of real-time classification on small-scale applications, with similar classification accuracy but much less parameters, compared with some state-of-the-art research results. In specific, the proposed deep CNN system incorporates a globally connected network topology with a generalized activation function. We then apply global average pooling (GAP) on the neurons of some hidden layers as well as the last convolution layers, of which the resulted vectors are concatenated together and fed into the softmax layer for classification. Henceforth, with only one classifier and one objective loss function for training, we shall enjoy the benefit of retaining rich information fused in the hidden layers while taking less parameters so that efficient information flow in both forward and backward stage is guaranteed,
and the overfitting risk is avoided.
Further, the proposed general activation function is composed of several of piecewise linear functions to approximate complex functions. We shall show that the proposed deep CNN architecture yields similar performance with much less parameters.

The contribution of this work is hence presented as follows.
\begin{itemize}
\item We present an architecture which makes full of use features learned at hidden layers, which avoids the gradient-vanishing problem to the most extent in backpropagation.
    The associated analysis on backward stage of the proposed architecture is also presented.
\item We define a generalized multi-piecewise ReLU activation function, which is able to approximate more complex and flexible functions and hence is shown to perform well in practice. The associated analysis on the forward pass as well as on the backpropagation are presented.
\item We also discuss the initialization of the associated parameters for further optimization of the presented work and the experiments show that our design achieves the state-of-the-art performance on several benchmark datasets in computer vision field.
\end{itemize}

The rest of the paper is organized as follows. In Section II and Section III,
the details of the proposed deep CNN architecture as well as the designed activation function are presented,
respectively,
including detailed analysis on both the forward phase and backward phase of
the presented architecture and the proposed activation function.
Section IV evaluates our design on several public datasets, such as MNIST \cite{LeCun-MNIST'98}, CIFAR-10/100 \cite{Hinton-Cifar'09}, SVHN \cite{Ng-SVHN'11}
as well as UCF Youtube Action Video Data Sets\cite{Action'UCFYoutube-1}\cite{Action'UCFYoutube-2}.
We conclude this paper in Section IV.

\section{Global-Connected Net (GC-Net)}
In this section, the proposed network architecture, namely Global-Connected Net (GC-Net), is discussed,
which is followed by the discussion of the proposed activation function in
Sec. \ref{sec:GReLU}.
\begin{figure}[!t]
\includegraphics[height=0.42\textwidth, width=0.42\textwidth, angle =90]{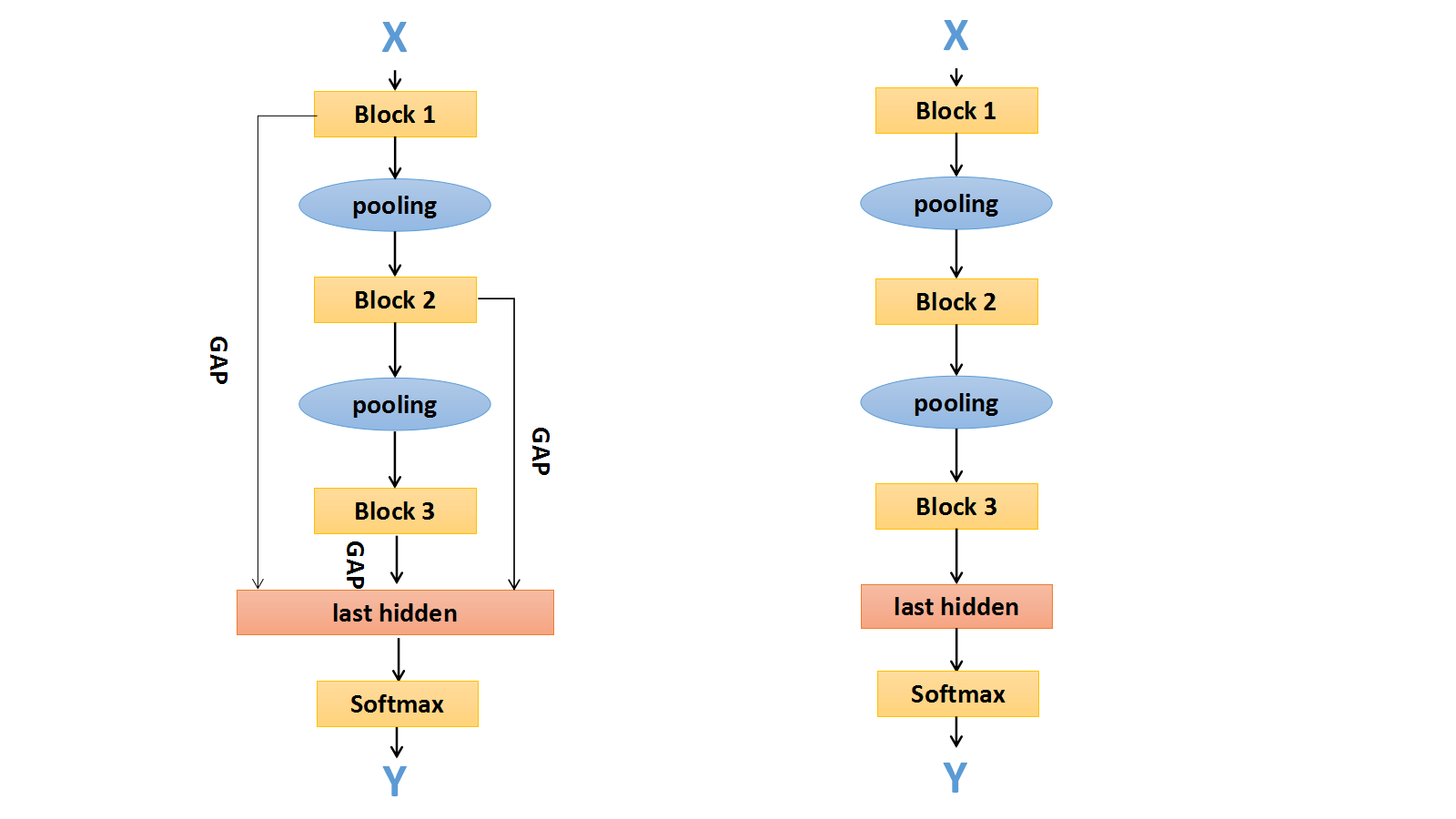}
\caption{Comparison of our proposed GlobeNet with conventional CNN architecture, where GAP stands for global average pooling.}
\label{fig:system}
\end{figure}

As shown in Fig. \ref{fig:system}, the proposed GC-Net, consists of $n$ blocks in total,
a fully-connected final hidden layer and a softmax classifier, where a block can have several convolutional layers, each followed by normalization layers and nonlinear activation layers. Max-pooling/Average pooling layers are
applied between connected blocks to reduce feature map sizes. The distinguished feature of the proposed GC-net network architecture from the conventional cascaded structure is that, we provide a direct connection between every block and the last hidden layer.
These connections in turn create a relatively larger vector full of rich features captured from all blocks,
which is fed as input into the last fully-connected hidden layer and then to the softmax clasifer
to obtain the classification probabilities in respective of labels.
In addition, to reduce the number of parameters in use, we only allow one
fully-connected layer to the final softmax classifier,
as more dense layers only has minimal performance improvement
while requires a lot of extra parameters.

In our GC-net design, to reduce the amount of parameters as well as computation burden,
we shall first apply global average pooling (GAP) to the output feature maps of all blocks and
then connects them with the last fully-connected hidden layer.
In this sense, we further flatten the neurons obtained from these blocks to obtain the
1-D vector for each blocks,
i.e., $\vec{\mathbf p}_i$ for block $i$ ($i=1,\cdots,N$) of length $m_i$. We then apply the concatenation operations on all of these 1-D vectors,
which hence resulted in a final 1-D vector consisting of neurons from these vectors,
i.e., $\vec{\mathbf p}=\overrightarrow{(\vec{\mathbf p}_1^T, \cdots,\vec{\mathbf p}_n^T)}^T$ with its length
defined as $m = \sum_{i=1}^N m_i$.
This resulted vector is then taken as the input to the last fully-connected hidden layer
before the softmax classifier for classification.
Therefore, to incorporate with this new feature vector,
a weight matrix
${\mathbf W}_{m \times s_c}=({\mathbf W}_{m_1 \times s_c}, \cdots, {\mathbf W}_{m_N \times s_c}) $
for the final fully-connected layer is required,
where $s_c$ is the number of classes of the corresponding dataset for recognition.
The final result fed into the softmax function hence is presented as,
\begin{align}
\vec{\mathbf c}^T = \vec{\mathbf p}{\mathbf W} =  \sum_{i=1}^N  \vec{\mathbf p}_i{\mathbf W}_i
\end{align}
i.e., $\vec{\mathbf c} ={\mathbf W}^T \vec{\mathbf p}^T$,
where ${\mathbf W_i}={\mathbf W}_{m_i \times s_c}$ for short.
$\vec{\mathbf c}^T$ is the input vector
into the softmax classifier, as well as the output of the fully-connected layer
with $\vec{\mathbf p}$ as input.

Therefore, in the back-propagation stage, defining
${dL}/{d \vec{\mathbf c}}$ is the gradient of the input fed to the
softmax classifier with respect to the loss function denoted by $L$
\footnote{Note that the gradient  ${dL}/{d \vec{\mathbf c}}$
can be readily derived from the standard back-propagation algorithm and
can be found in the textbooks in the literature
and hence is omitted.},
the gradient of the concatenated vector is then given by,
\begin{align}
\frac {d L} {d \vec{\mathbf p}}
=  \frac {d L} {d \vec{\mathbf c}}\frac {d \vec{\mathbf c}} {d \vec{\mathbf p}}
= {\mathbf W}^T \frac{d L} {d \vec{\mathbf c}}
= (\frac {d L} {d \vec{\mathbf p}_1},\cdots, \frac {d L}{d \vec{\mathbf p}_n})
\end{align}
Therefore, for the resulted vector $\vec{\mathbf p}_i$ after pooling from the output of block $i$,
we obtain its gradient ${d L} / {d \vec{\mathbf p}_i}$ directly from the softmax classifier.

Further, taking the cascaded back propagation process into account, except block $n$,
all other blocks will also receive the gradients
from its following block in the backward pass.
Let us define the output of block $i$ as ${\mathbf B}_i$ and the final gradient of the output of block $i$
with respect to the loss function as $\frac{d L}{d {\mathbf B}_i}$.
Then, taking both gradients combing from the final layer and the adjacent block of the cascaded
structure into account, the derivation of $\frac{d L}{d \vec{\mathbf B}_i}$
is hence summarized in Lemma \ref{lem:gradient},
\begin{lemma} \label{lem:gradient}
The full gradient to the output of block $i$ ($i<n$) with respect to the loss function is given by,
\begin{align}
\frac{d L}{d {\mathbf B}_i} =
\frac{d L} {d \vec{\mathbf p}_i} \frac{d {\mathbf B}_i}{d \vec{\mathbf p}_i}
+ \left(\frac{d L}{d \vec{\mathbf p}_n}\frac{d \vec{\mathbf B}_n}{d \vec{\mathbf p}_n} \right)
\sum_{j=i}^{n-1}
\frac{d {\mathbf B}_{j+1}} {d {\mathbf B}_j}
\end{align}
where $\frac{d{\mathbf B}_{j+1}} {d {\mathbf B}_j}$ is defined as the
gradient for the cascaded structure from block $j+1$ back-propagated to block of $j$
and $\frac{d {\mathbf B}_i}{d \vec{\mathbf p}_i}$ is the gradient of output of block $i$
${\mathbf B}_i$ with respect to its pooled vector $\vec{\mathbf p}_i$.
\end{lemma}

The proof of Lemma \ref{lem:gradient} is straightforward from the chains rule in differentiation
and hence is omitted.
As observed in Lemma \ref{lem:gradient},
each hidden block can receive gradients
benefitted from its direct connection with the last fully connected layer. Interestingly,
the earlier hidden blocks can even receive more gradients, as it not only receive the gradients directly from the last layer,
back-propagated from the standard cascaded structure,
but also those gradients back-propagated from
the following hidden blocks with respect to their direct connection with the final layer.
Therefore, the gradient-vanishing problem is expected to be mitigated to some extent.
In this sense, the features generated in the hidden layer neurons are well exploited and relayed for classification.

It is noted that our design differs from all the reported research in the literature as it builds connections among blocks, instead of only within blocks, such as ResNet \cite{He'ResNet'16} and Dense-connected nets \cite{Densely-connected'16}. Our design is also different from the deep-supervised nets in \cite{Lee'DeepSuper'14} which connects every hidden layer with an independent auxiliary classifier (and not the final layer) for regularization but the parameters with these auxiliary classifiers are not used in the inference stage, hence results in inefficiency of parameters utilization.
In our design, in contrast to the deep-supervised net \cite{Lee'DeepSuper'14},
each block is allowed to connect with the last hidden layer that connects with only one final softmax layer for classification,
for both the training and inference stages.
All of the designed parameters are hence efficiently utilized to the most extent,
especially in the inference stage, compared with \cite{Lee'DeepSuper'14}.

Note also that by employing global average pooling (i.e., using a large kernel size for pooling)
prior to the global connection in our design,
the number of resulted features from all blocks is greatly reduced,
which hence significantly simplifies our structure and
make the extra number of parameters brought by this design minimal. Further,
this does not affect the depth of the neural network, hence has
negligible impact on the overall computation overhead.
It is further emphasized that, in back-propagation stage, each block can receive gradients coming from both the
cascaded structure and directly from the generated 1-D vector as well, thanks to the newly
added connections between each block and the final hidden layer. Thus, the weights of the
hidden layer will be better tuned, leading to higher classification performance.

\begin{figure}[!t]
\includegraphics[height=0.42\textwidth, width=0.42\textwidth]{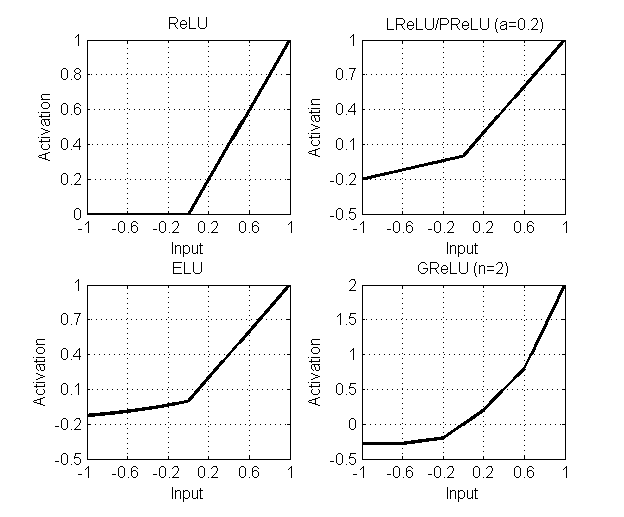}
\caption{Description of GReLU with other popular nonlinear activation functions.
In GReLU, we set
the ranges of sections are $\left((-\infty,-0.6),(-0.6,-0.2),(-0.2,0.2),(0.2,0.6),(0.6,\infty)\right)$
and the corresponding slopes for these sections are $(0.01, 0.2, 1, 1.5, 3)$, respectively.}
\label{fig:grelu}
\end{figure}

\section{Generalized ReLU Activation}\label{sec:GReLU}
To collaborate with GC-Net, a new type of nonlinear activation function is proposed
and the details are presented as follows.

\subsection{Definition and Forward Phase of GReLU}
As shown in Fig. \ref{fig:grelu}, the Generalized Multi-Piecewise ReLU, termed GReLU,
is defined as a combination of many
piecewise linear functions as presented in (\ref{eq:Grelu}) on top of next page.
\begin{figure*}
\centering
\begin{align}
y(x) = \left
\{ \begin{array}{ll}
         l_1+\sum_{i=1}^{n-1} k_i(l_{i+1}-l_i)+k_n(x-l_n), & \mbox{if $x \in [l_n,\infty)$};\\
         \vdots\\
         l_1+k_1(x-l_1),  & \mbox{if $x \in [l_1,l_2)$};\\
         x & \mbox{if $x \in [l_{-1},l_1)$};\\
         l_{-1}+k_{-1}(x-l_{-1}) , & \mbox{if $x \in [l_{-2},l_{-1})$};\\
         \vdots\\
         l_{-1}+\sum_{i=1}^{n-1}k_{-i}(l_{-(i+1)}-l_{-i})+k_{-n}(x-l_{-n}), & \mbox{if $x \in (-\infty,l_{-n})$}.
\end{array}
\right. \label{eq:Grelu}
\end{align}
\end{figure*}
As defined in (\ref{eq:Grelu}), if the inputs fall into the center range of $(l_{-1},l_1)$,
the slope is set to be unity and the bias is set to be zero, i.e.,
identity mapping is applied.
Otherwise, when the inputs are larger than $l_1$, i.e., they fall into one of the ranges on the positive direction in $\{(l_1,l_2),\cdots,(l_{n-1},l_n),(l_n,\infty)\}$,
and we assign slopes ($k_1,\cdots,k_n)$ to those ranges, respectively.
The bias can then be readily calculated
from the multi-piecewise linear structure of the designed function.
Similarly, if the inputs fall into one of the ranges on the negative direction in $\{(l_{-1},l_{-2}),\cdots,(l_{-(n-1)},l_{-n}),(l_{-n},-\infty)\}$
we assign $(l_{-1},\cdots,l_{-(n-1},l_{-n})$ to all of those ranges, respectively.
By doing so, the useful features
learned from linear mappings like convolution and fully-connected operations
are hence boosted through the designed GReLU activation function.

To fully exploit the designed multi-piecewise linear activation function,
both the endpoints $l_i$ and slopes $k_i$ ($i=-n,\cdots,-1,1,\cdots,n$) are set to be
learnable parameters, and for simplicity and computation efficiency we restrict on
channel-shared learning for the designed GReLU activation functions.
Further, we do not impose constraints on the leftmost and rightmost points,
which are then learned freely while the training goes on.

Therefore, for each activation layer, GRuLU only has $4n$ ($n$ is the number of ranges on both directions) learnable parameters,
wherein $2n$ accounts for the endpoints and another $2n$ for the slopes of the piecewise linear functions, which is definitely negligible compared with millions of parameters in current popular deep CNN models.
For example, GoogleNet has $5$ million parameters and 22 layers. It is evident that,
with increased $n$, GReLU can approximate complex functions even better
at the cost of extra computation resources consumed,
but in practice even a small $n$ ($n=2$) suffices for image/vedeo classification tasks.

\subsection{Relation to Other Activation Functions}
It is readily observed that GReLU is an generalization of its prior counterparts.
For example, setting the slopes of all sections in the positive range to be unity and
that of the sections in the negative direction a shared value, it degenerates into
leaky ReLU if the update of parameters is not allowed, and PReLU otherwise.
Further, by setting the slopes of the negative side to be zero, it is degenerated into ReLU
function. In this sense, GReLU is an extension to these functions,
and hence has the potential of more flexible feature learning capabilities and
should perform better than its counterparts.

\subsection{Backward Phase of GReLU}
Regarding the training of GReLU, the gradient
descent algorithm for back-propagation is applied.
The derivatives of the activation function with respect to the input as well as the learnable parameters are hence given in
(\ref{eq:deri_x})-(\ref{eq:deri_l}) (where (\ref{eq:deri_k}) is presented on top of next page)
as follows.
\begin{align}
 \pdv{y(x)}{x} = \left
\{ \begin{array}{ll}
        k_n,  & \mbox{if $x \in [l_n,\infty)$};\\
         \vdots\\
        k_{1},   & \mbox{if $x \in [l_1,l_2)$};\\
         1, & \mbox{if $x \in [l_{-1},l_1)$};\\
         k_{-1},  & \mbox{if $x \in [l_{-2},l_{-1})$};\\
         \vdots\\
         k_{-n}, & \mbox{if $x \in (-\infty,l_{-n})$}.
\end{array}
\right.   \label{eq:deri_x}
\end{align}
where the derivative to the input is simply
the slope of the associated linear mapping when the input falls in its range.
\begin{figure*}
\centering
\begin{align}
\pdv{y(x)}{k_i} = \left
\{ \begin{array}{ll}
        (l_{i+1}-l_i) I\{x>l_{i+1}\}+(x-l_i)I\{l_i<x \leq l_{i+1}\},  & \quad \mbox{if $i\in\{1,\cdots,n-1\}$};\\
        (x-l_i)I\{x>l_i\}, & \quad \mbox{if $i=n$}; \\
        (x-l_i)I\{x \leq l_i\}, & \quad \mbox{if $i=-n$}; \\
        (l_{i-1}-l_i) I\{x<l_{i-1}\}+(x-l_i)I\{l_{i-1} <x \leq l_{i}\}, & \quad \mbox{if $i\in\{-n+1,\cdots,-1\}$}.
\end{array}
\right.   \label{eq:deri_k}
\end{align}
\end{figure*}
\begin{align}
\pdv{y(x)}{l_i} = \left
\{ \begin{array}{ll}
        (k_{i-1}-k_i) I\{x>l_i \},  & \quad \mbox{if $i>1$};\\
        (1-k_1) I\{x>l_1 \},   & \quad \mbox{if $i = 1$};\\
          (1-k_{-1}) I\{x<=l_{-1} \}, & \quad \mbox{if $i = -1$};\\
          (k_{i+1}-k_i) I\{x<=l_i \}, & \quad \mbox{if $i<-1$}.
\end{array}
\right.  \label{eq:deri_l}
\end{align}
where $I\{ \cdot \}$ is an indication function returning unity when the event $\{ \cdot \}$ happens
and zero otherwise.

The back-propagation update rule for the parameters of GReLU activation function can be derived by chain rule as follows,
\begin{align}
\pdv{L}{o_i} = \sum_{j}\pdv{L}{y_j}\pdv{y_j}{o_i} \label{eq:BP}
\end{align}
where $L$ is the loss function, $y_j$ is the output of the activation function,
and $o_i \in\{k_i,l_i\}$ is the learnable parameters of GReLU. Note that the summation is applied in all positions and across all feature maps for the activated output of the current layer, as the parameters are channel-shared.
$\pdv{L}{y_j}$ is defined as the derivative of the activated GReLU output back-propagated from the loss function through its upper layers.
Therefore, the simple update rule for the learnable parameters of GReLU activation function is
\begin{align}
o_i \leftarrow o_i - \alpha \pdv{L}{o_i}
\end{align}
where $\alpha$ is the learning rate. The weight decay (e.g., $L2$ regularization) is not taken into account in updating these parameters.

\subsection{Benefits of GReLU}
From the discussion above, it is therefore found out that designing
GReLU as a multi-piecewise linear functions
has several benefits, compared to its counterparts.
One is that it is enabled to approximate complex functions
whether they are convex functions or not, while most of activation functions however cannot.
This demonstrates its stronger capability in feature learning.
Further, since it only employs linear mappings in different ranges along the dimension,
it inherits
the advantage of the non-saturate functions, i.e., the gradient vanishing/exploding effect is mitigated to the most extent.
We shall discuss its effect further in the experiment part.

\section{Experiments and Analysis}
\subsection{Overall Setting}
The following public datasets with different scales, MNIST\cite{LeCun-MNIST'98}, CIFAR10, CIFAR100\cite{Hinton-Cifar'09}, SVHN\cite{Ng-SVHN'11}, and UCF YouTube Action Video datasets,
are employed to test the proposed GC-Net and GReLU. Experiments are firstly conducted on small neural nets using the small dataset MNIST and compare the resultant performance with that by the traditional CNN schemes. Then we move to a larger CNN for performance comparison with other state-of-the-art models, such as stochastic pooling \cite{Stochastic-pooling'13}, NIN \cite{M'13-NIN} and Maxout\cite{Goodfellow-Maxout'13}, for all of these datasets.
Due to the complexity of GReLU, we freeze the learning of slopes and endpoints
in the first few epochs treating it like a Leaky ReLU function,
and then starts to learn them in the epochs thereafter. Our experiments are implemented in
PYTORCH with one Nvidia GeForce GTX 1080.
\footnote{
The hyperparameters used in all the experiments are not extensively optimized due to the computation resources constraint
while
the slopes and end-points of GReLU are manually initialized. Better results are hence expected with extensive search for these parameters.}.

\subsection{MNIST On SmallNet}
The MNIST digit dataset \cite{LeCun-MNIST'98} contains $70,000$ $28 \times 28$
gray scale images of numerical digits from $0$ to $9$. The dataset
is divided into the training set with $60,000$ images
and the test set with $10,000$ images.

In this SmallNet experiment, MNIST is used for performance comparison between our model with conventional ones.
The proposed GReLU activated GC-Net is composed of 3 convolution layers with small $3 \times 3$ filters and only $16$, $16$ and $32$ feature maps, respectively. The $2 \times 2$ max pooling layer with a stride of $2 \times 2$
was applied after both of the first two convolution layers.
GAP is applied to the output of each convolution layer and the collected averaged features are
fed as input to the softmax layer for classification. The total number of parameters amounts to be only around $8.3K$. For a fair comparison,
we also examined the dataset using a $3$-convolution-layer CNN with ReLU activation, with $16$, $16$ and $36$ feature maps equipped in the three convolutional layers, respectively. Therefore, both tested networks use a similar amount of parameters (if not the same).

\begin{figure}
\includegraphics[height=0.44\textwidth, width=0.44\textwidth]{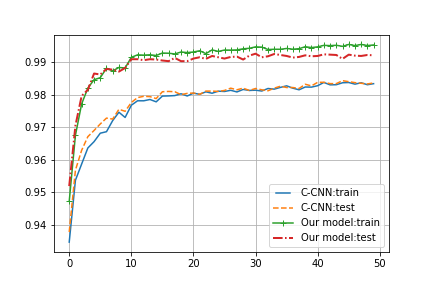}
\caption{Comparison result of our designed model with conventional CNN (C-CNN) with ReLU activation.}
\label{fig:MNIST}
\end{figure}

In MNIST, neither preprocessing nor data augmentation was performed on this dataset, except we re-scale the pixel values to be within ($-1,1$) range.
The experiment result in Fig. \ref{fig:MNIST} shows that the proposed GReLU activated GC-Net achieves an error rate no larger than $0.78\%$ compared with $1.7\%$ by the conventional CNN, which is over $50 \%$ of improvement in accuracy, after a run of 50 epochs.
It is also observed that the proposed architecture tends to converge fast, compared with its conventional counterpart.
In fact, for our model,
test accuracy exceeds below $1\%$ error rate only starting from epoch $10$,
while the conventional net reaches similar performance
only after epoch $15$.

We have also conducted other experiment on the MNIST dataset to further verify its performance with relatively more complex models.
Different from the previous one, we kept all the schemes to achieve similar error rates while observing the required number of trained parameters. Again, we used a network with three convolutional layers by keeping all convolutional layers with $64$ feature maps and $3\times3$ filters. The experiment results are shown in Table \ref{tab:MNIST}, where the proposed GC-Net with GReLU yields a similar error rate (i.e., $0.42\%$ versus $0.47\%$) while taking only $25\%$ of the total trained parameters by its counterparts. The results of the two experiments on MNIST clearly demonstrated the superiority of the proposed GReLU activated GC-Net over the traditional CNN schemes. Further, with roughly $0.20$M parameters, a relatively larger network with our framework achieves the state-of-art accuracy performance, i.e., $0.28\%$ error rate,  while its benchmark counterparts, DSN, achieves $0.39\%$ error rate with a total of $0.35$M parameters.

\begin{table}
  \centering
  \caption{Error rates on MNIST without data augmentation.}
  \label{tab:MNIST}
  \begin{tabular}{ccc}
   \hline
    Model & No. of Param.(MB) & Error Rates\\
    \hline
    Stochastic Pooling  & $0.22$M                    & $0.47\%$  \\
    Maxout                   & $0.42M$       & $0.47\%$  \\
    DSN+softmax                   & $0.35M$              & $0.51\%$ \\
    DSN+SVM           & $0.35M$              & $0.39\%$ \\
    NIN + ReLU         & $0.35M$                &  $ 0.47\%$ \\
    NIN + SReLU      & $0.35M+5.68K$   &  $0.35\%$  \\
    GReLU-GC-Net        &   $0.078M$                    &  ${\bf 0.42}\%$ \\
    GReLU-GC-Net        &   $0.22M$                    &  ${\bf 0.27}\%$ \\
    \hline
  \end{tabular}
\end{table}

\subsection{CIFAR10}
The CIFAR-10 dataset contain $60,000$ natural color (RGB) images
with a size of $32 \times 32$ in $10$ general object classes.
The dataset is divided into $50,000$ training images and $10,000$
testing images. All of our experiments are implemented
without data augmentation and we employ the same preprocessing strategy in \cite{M'13-NIN}.
The comparison results
of the proposed GReLU activated GC-Net to the reported methods in the literature on this dataset, including stochastic pooling \cite{Stochastic-pooling'13}, maxout \cite{Goodfellow-Maxout'13}, prob maxout \cite{Springenberg-prob-maxout'13}, NIN \cite{M'13-NIN}, are given in Table. \ref{tab:cifar10}.
It is observed that our method achieves comparable performance while taking greatly reduced number of parameters employed in other models. Interestingly, one of our shallow model with only $0.092M$ parameters in $3$ convolution layers achieves comparable performance with convolution kernel method in \cite{Conv-kernel'14}.
For the experiments with 6 convolution layers, with roughly $0.61$M parameters,
our model achieves comparable performance in contrast to Maxout with $5$M parameters.
Actually, compared with NIN consisting of $9$ convolution layers and roughly $1$M parameters,
our model achieves competitive performance,
only in a 6-convolution-layer shallow architecture
with roughly $60\%$ of parameters of it.
These results hence well demonstrate the advantage of our proposed GReLU activated GC-Net method, which accomplishes similar performance with less parameters and
a shallower structure (less convolution layers required), and hence is appropriate for memory-efficient and computation-efficient scenarios, such as mobile applications.
\begin{table}
  \centering
  \caption{Error rates on CIFAR-10 without data augmentation.}
  \label{tab:cifar10}
  \begin{tabular}{lll}
  \hline
    Model & No. of Param.(MB) & Error Rates\\
    \hline
    Conv kernel \cite{Conv-kernel'14} &-        & $17.82\%$ \\
    Stochastic pooling  & -        & $15.13\%$ \\
    ResNet \cite{Stochastic-Resnet'16} (110 layers)  &$1.7M$  & $13.63\%$\\
    ResNet \cite{Stochastic-Resnet'16} (1001 layers)  &$10.2M$  & $10.56\%$\\
    Maxout                   & $>5M$       & $11.68\%$  \\
    Prob Maxout                   & $>5M$       & $11.35\%$  \\
    DSN (9 conv layers)                         &$0.97M$                &$9.78\%$   \\
    NIN (9 conv layers)                   &    $0.97M$                &  $ 10.41\%$ \\
    Ours  (3 conv layers)             &   $0.092M$                    &  $17.23\%$ \\
    Ours (6 conv layers)             &    ${\bf 0.11M}$            &        ${\bf 12.55\%}$ \\
    Ours (6 conv layers)              &   $0.61M$                    &  $10.39\%$ \\
    Ours (8 conv layers)               &   $0.91M$                    &  $9.38\%$ \\
    \hline
  \end{tabular}
\end{table}

\subsection{CIFAR100}
The CIFAR-100 dataset also contain $60,000$ natural color (RGB) images
with a size of $32 \times 32$ but in 100 general object classes.
The dataset is divided into $50,000$ training images and $10,000$
testing images.
Our experiments on this datasset are implemented
without data augmentation and we employ the same preprocessing strategy in \cite{M'13-NIN}.
The comparison results
of our model to the reported methods in the literature
on this dataset, are given in Table. \ref{tab:cifar100}.
It is observed that our method achieves comparable performance while taking
greatly reduced number of parameters employed in other models.
As observed in Table. \ref{tab:cifar100}, one of our shallow model
with only $0.16M$ parameters and $3$ convolution layers,
achieves comparable performance with deep ResNet in \cite{Stochastic-Resnet'16} of $1.6$M parameters.
In the experiments with $6$ convolution layers, it is observed that,
with roughly $10\%$ of parameters in Maxout,
our model achieves comparable performance in contrast to it.
In addition, with roughly $60\%$ of parameters of NIN,
our model accomplishes competitive (or even slightly higher) performance than it,
which however consists of $9$ convolution layers ($3$ layer deeper than the compared model).
This hence validates the powerful feature learning capabilities of our designed GC-net
with GReLU activations. In such way, we can achieve similar performance with shallower structure and
less parameters.

\begin{table}
  \centering
  \caption{Error rates on CIFAR-100 without data augmentation.}
  \label{tab:cifar100}
  \begin{tabular}{lll}
   \hline
    Model & No. of Param.(MB) & Error Rates\\
    \hline
    ResNet \cite{Stochastic-Resnet'16}  &$1.7M$  & $44.74\%$\\
    Stochastic pooling  &   -                &$42.51\%$ \\
    Maxout                   & $>5M$       & $38.57\%$  \\
    Prob Maxout                   & $>5M$       & $38.14\%$  \\
    DSN                        &$1M$                 &$34.57\%$ \\
    NIN  (9 conv layers)                   &    $1M$                &  $ 35.68\%$ \\
    Ours (3 conv layers)               &   $0.16M$                    &  $44.79\%$ \\
    Ours (6 conv layers)               &   $0.62M$                    &  $35.59\%$ \\
    Ours (8 conv layers)               &   $0.95M$                    &  $33.87\%$ \\
    \hline
  \end{tabular}
\end{table}

\subsection{Street View House Numbers (SVHN)}
The SVHN Data Set contains $630,420$ RGB images of house numbers,
collected by Google Street View. The dataset comes in two formats and we only
consider the second format, with all images being of
size $32 \times 32$ and the task is to classify the digit in
the center of the image, however possibly some digits may appear
beside it but are considered noise and ignored.
This dataset is splitted into three subset,
i.e., extra set, training set, and test set, and each with $531,131$, $73,257$,
and $26,032$ images, respectively,
where the extra set is a less difficult set used to be extra training set.
Compared with MNIST, it is a much more challenging digit dataset due to its
large color and illumination variations.

In SVHN, in data preprocessing, we simply re-scale the pixel values to be within ($-1,1$) range, identical to that imposed on MNIST.
It is noted that for other methods, local constrast method is employed for data preprocessing.
Even so, it is observed that, our models are quite competitive compared with
other baseline models. For example, one model with only $6$ convolution layers and $0.61$M parameters,
it achieved roughly the same performance with NIN, which consists of $9$
convolution layers and around $2$M parameters.
Further, for a deeper models with $9$
layers and $0.90$M parameters, we achieve the state of art performance,
which validates the powerful feature learning capabilities of the designed architecture.

\begin{table}
  \centering
  \caption{Error rates on SVHN.}
  \label{tab:cifar100}
  \begin{tabular}{lll}
   \hline
    Model & No. of Param.(MB) & Error Rates\\
    \hline
    Stochastic pooling  &   -                &$2.80\%$ \\
    Maxout                   & $>5M$       & $2.47\%$  \\
    Prob Maxout                   & $>5M$       & $2.39\%$  \\
    DSN                        &$1.98M$                 &$1.92\%$ \\
    NIN  (9 conv layers)                   &    $1.98M$                &  $ 2.35\%$ \\
    Ours (6 conv layers)               &   $0.61M$                    &  $2.35\%$ \\
    Ours (8 conv layers)               &   $0.90M$                    &  $2.10\%$ \\
    \hline
  \end{tabular}
\end{table}

\subsection{UCF YouTube Action Video Dataset}
The UCF YouTube Action Video Dataset is a popular video dataset for action recognition.
It is consisted of approximately 1168 videos in total and contains $11$ action categories, including: basketball shooting, biking/cycling,
diving, golf swinging, horse back riding, soccer juggling, swinging, tennis swinging,
trampoline jumping, volleyball spiking, and walking with a dog.
For each category, the videos are grouped into 25 groups with over 4 action clips in it.
The video clips belongs to the same group may share some common characteristics,
such as the same actor, similar background, similar viewpoint, and so on.
The dataset is split into training set and test set, each with $1,291$ and $306$ samples, respectively.
It is noted that UCF YouTube Action Video Dataset is quite challenging due to large variations in camera motion,
object appearance and pose, object scale, viewpoint, cluttered background,
illumination conditions, etc. Similar to \cite{Youtube-benchmark},
for each video in this dataset,  we select $16$ non-overlapping frames clips.
However, due to the limitation of GPU memory,
we simply resize each frame into size $36 \times 36$ and then crop the centered
$32 \times 32$ for training to guarantee that
a batch of 16 samples is fit for $8$G memory of GTX$1080$.
Due to the down-sampling implemented on original frames, some performance degradation is expected.
Further, we only allow convolution across space within each frame
but not across frames over time epochs for this video classification task.
This however will definitely have some negative impact on performance,
as in fact we do not try to exploit the temporal relationship of frames to the most extent.
Even with such simple setup, our designed neural network is capable of achieving higher performance,
than the benchmark method using hybrid features in \cite{Action'UCFYoutube-1}.

\begin{table}
  \centering
  \caption{Error rates on UCF Youtube Action Video Dataset.}
  \label{tab:cifar100}
  \begin{tabular}{lll}
   \hline
    Model & No. of Param.(MB) & Error Rates\\
    \hline
    \cite{Action'UCFYoutube-1}: static features  &-     &  $ 63.1\%$ \\
    \cite{Action'UCFYoutube-1}: motion features  &-     &  $65.4\%$ \\
    \cite{Action'UCFYoutube-1}: hybrid features  &-     &  $71.2\%$ \\
    Ours               &  -                    &  $72.6\%$ \\
    \hline
  \end{tabular}
\end{table}

\section{Conclusion}
In this work, we have designed an architecture making better use of the hidden layer features, to alleviate the gradient-vanishing problem. Further, a generalized linear rectifier activation function was proposed to boost the performance. The combination of the two designs is demonstrated to achieve state of art performance in several object recognition and video action recognition
benchmark tasks, including MNIST, CIFAR-$10/100$, SVHN and UCF YouTube Action video datasets,
with greatly reduced amount of parameters and even shallower structure. Henceforth, our design can be employed in small-scale real-time application scenarios, as it requires less parameters and shallower network structure whereas achieving matching/close performance with state-of-the-art models. In our future work, we shall further tune the network performance and employ it to deeper neural networks
in other interesting machine learning tasks such as autonomous driving.

\end{document}